\documentclass{article}

\usepackage[usenames,dvipsnames,svgnames,table]{xcolor}
\usepackage[breaklinks=true,colorlinks,bookmarks=false,citecolor=MidnightBlue,linkcolor=BrickRed]{hyperref}
\usepackage{nips13submit_e,times}
\usepackage[round]{natbib}
\usepackage{subfigure}
\usepackage{amsmath}
\usepackage{graphicx}

\graphicspath{{images/}}
\DeclareGraphicsExtensions{.pdf,.png,.jpeg}

\newcommand{\eg}{\emph{e.g.},}

\newcommand{\ie}{\emph{i.e.},}

\newcommand{\etc}{\emph{etc.}}

\DeclareUrlCommand\emailurl{}
\newcommand{\email}[1]{\href{mailto:#1}{\emailurl{#1}}}

\newcommand{\red}[1]{{\color{red}#1}}
\newcommand{\green}[1]{{\color{green}#1}}

\title{Discriminative Training: Learning to Describe Video with Sentences,
  from Video Described with Sentences}

\author{
Haonan Yu\thanks{School of Electrical and Computer
  Engineering, Purdue University, West Lafayette IN 47907-2035}\\
\email{haonan@haonanyu.com}\\
\And
Jeffrey Mark Siskind\footnotemark[1]\\
\email{qobi@purdue.edu}
}

\nipsfinalcopy %

\begin{document}

\maketitle

\begin{abstract}
We present a method for learning word meanings from complex and realistic video
clips by discriminatively training (DT) positive sentential labels against
negative ones, and then use the trained word models to generate sentential
descriptions for new video.
This new work is inspired by recent work which adopts a maximum likelihood (ML)
framework to address the same problem using only positive sentential labels.
The new method, like the ML-based one, is able to automatically determine which
words in the sentence correspond to which concepts in the video (\ie\ ground
words to meanings) in a weakly supervised fashion.
While both DT and ML yield comparable results with sufficient training data, DT
outperforms ML significantly with smaller training sets because it can exploit
negative training labels to better constrain the learning problem.
\end{abstract}

\section{Introduction}
Generating linguistic description of visual data is an important topic at the
intersection of computer vision, machine learning, and natural-language
processing.
While most prior work focuses on describing static images
\citep{Yao2010,Kulkarni2011,Ordonez2011,Kuznetsova2012}, little work focuses
on describing video data.
\citet{Kojima2002} established the correspondence between linguistic concepts
and semantic features extracted from video to produce case frames which were
then translated into textual descriptions.
\citet{Lee2008} used a stochastic context free grammar (SCFG) to infer events
from video images parsed into scene elements.
Text sentences were then generated by a simplified head-driven phrase structure
grammar (HPSG) based on the output of the event inference engine.
\citet{Khan2012} extracted high level features (\eg\ semantic keywords) from
video and then implemented a template filling approach for sentence generation.
\citet{Barbu2012a} used a detection-based tracker to track object motion,
hidden Markov models (HMM) to classify the object motion into verbs, and
templates to generate sentences from the verbs, detected object classes, and
track properties.
\cite{Krishnamoorthy2013} combined object and activity detectors with knowledge
automatically mined from web-scale text corpora to select the most likely
subject-verb-object (SVO) triplet.
This triplet was then expanded into a sentence by filling a template.
These approaches use a common strategy for generating descriptions, namely
mosaicing together different parts of a sentence.
They often employ different mechanisms for different parts of speech; while
verbs are often represented by learned event models or grammars, \emph{ad hoc}
hand-coded knowledge is often used to represent other word types such as
prepositions and adverbs.
Such separate handling of different parts of speech is unprincipled and
requires greater effort to craft a system by hand or label larger amounts of
training data.

\citet{Barbu2012b} presented a method that combines detection-based tracking
with event recognition based on HMMs.
They formed a factorial HMM with the cross product of the \citet{Viterbi1967}
lattices for both the detection-based tracking process and the event
recognition HMM, finding the maximum \emph{a posteriori} probability (MAP)
estimate of a track that both exhibits temporal coherency as required by
detection-based tracking and the motion profile described by the HMM.\@
A companion submission extends this work to support multiple object tracks
mutually constrained by multiple hand-coded HMMs denoting the semantic meaning
representations for different words in a sentence, each applied to a subset of
the tracks.
They called this the \emph{sentence tracker}.
\citet{Yu2013} built upon this work to train the HMMs from a corpus of video
clips paired with sentential descriptions.
Their method was able to learn word meanings in a weakly supervised fashion:
while the video clips were paired with multi-word sentential labels, the
learner was \emph{not} provided the mapping from the individual words to the
corresponding semantic notions in the video.
Their approach was motivated by the hypothesis that children acquire language
through \emph{cross-situational} learning \citep{Siskind1996}.
While there exist many potential word-to-meaning mappings that are consistent
with a single video-sentence training sample, fewer such mappings will be
consistent as the number of training samples increases.
This yields a constraint satisfaction problem (CSP), where each training sample
acts as a constraint on the mutual meanings of the words in that sentence and
information learned about the meaning of one word flows to other words in that
sentence and on to other words in other sentences.
This cross-situational aspect of their algorithm allowed it to correctly
learn the meanings of all words in all sentences that appeared in their
training corpus.
After this, they were able to decide whether a video depicted a new sentence
by thresholding the video-sentence score computed with the learned word HMMs.

However, there is a potential issue in their maximum likelihood (ML)
formulation.
It works well when sufficient training data is provided to constrain the
problem so that only a single word-to-meaning mapping is consistent with the
training set.
When multiple word-to-meaning mappings are consistent, it is possible that
an incorrect mapping yields higher likelihood.
Having only a small number of sentential labels for a small number of video
clips may yield insufficient constraint on the learning problem.
The essence of this paper is to remedy this problem by automatically generating
negative training sentences for a video, thus increasing the degree of
constraint on the consistent word-to-meaning mappings without requiring
additional training video clips.
These automatically generated negative (training) sentences describe what did
\emph{not} occur in a video clip in contrast to the manually specified
positive (training) sentences.
The hypothesis is that such information will yield a more constrained learning
problem with the same amount of video data.
The contribution of this paper is to give a discriminative training (DT)
formulation for training positive sentences against negative ones.
This strictly generalizes the ML-based method, as ML is equivalent to DT with
an empty negative training set.

The remainder of the paper is organized as follows.
Section~\ref{section:background} reviews the ML formulation which serves as the
basis of the work in this paper.
Section~\ref{section:dt} describes the DT formulation and learning algorithm.
Section~\ref{section:training-regimen} proposes a two-phase regimen
combining ML and DT for training.
We demonstrate the advantage of DT over ML on an example in
Section~\ref{section:experiment}.
Finally, we conclude with a discussion in Section~\ref{section:conclusion}.

\section{Background}
\label{section:background}
Table~~\ref{tab:notation} summarizes our notation, which extends that in
\citet{Yu2013}.
The training set contains training samples, each pairing a video clip with a
sentence.
The method starts by processing each video clip with an object detector to
yield a number of detections for each object class in each frame.
To compensate for false negatives in object detection, detections in each frame
are \emph{overgenerated}.
Consider a track~$\tau_u$ to be a sequence of detections, one in each frame,
constrained to be of the same class.
Conceptually, there are exponentially many possible tracks, though we do not
need to explicitly enumerate such, instead implicitly quantifying over such by
way of the \citet{Viterbi1967} algorithm.

The method is also given the \emph{argument-to-participant mapping} for each
sentence.
For example, a sentence like \emph{The person to the left of the backpack
  approached the trash-can} would be represented as a conjunction
\begin{equation*}
\textit{person}(p_0)\wedge
\textit{to-the-left-of}(p_0,p_1)\wedge
\textit{backpack}(p_1)\wedge
\textit{approached}(p_0,p_2)\wedge
\textit{trash-can}(p_2)
\end{equation*}
over the three participants~$p_0$, $p_1$, and~$p_2$.
This could be done in the abstract, without reference to a particular video,
and can be determined by parsing the sentence with a known grammar and a lexicon
with known arity for each lexical entry.
Each lexical entry is associated with an HMM that models its semantic meaning.
HMMs associated with entries of the same part-of-speech have the same model
configuration (\ie\ number of states, parametric form of output distribution,
\etc).

\begin{table}
  \begin{center}
    \begin{tabular}{@{}l@{\hspace*{3pt}}l|l@{\hspace*{3pt}}l@{}}
      $M$ & number of entries in the lexicon &
      $C_m$ & part-of-speech of lexical entry $m$\\
      $R$ & number of training samples &
      $D_r$ & video clip in training sample~$r$\\
      $S_r$ & sentence in training sample~$r$ &
      $L_r$ & number of words in sentence~$S_r$\\
      $S_{r,l}$ & $l$th word in sentence~$S_r$ &
      $T_r$ & number of frames in video~$D_r$\\
      \red{$j^t_r$} & sequence of detection indices in &
      \red{$j_r$} & $(j^1_r,\ldots,j^{T_r}_r)$\\
      & frame~$t$ of video~$D_r$, one index per track\\
      \red{$q^t_{r,l}$} & state of the HMM for word~$l$ in &
      \red{$q_{r,l}$} & $(q^1_{r,l},\ldots,q^{T_r}_{r,l})$\\
      & sentence~$S_r$ at frame~$t$\\
      \red{$q_r$} & $(q_{r,1},\ldots,q_{r,L_r})$ &
      $I_c$ & number of states in the HMM for\\
      &&& part-of-speech~$c$\\
      $N_c$ & length of the feature vector for &
      \red{$x^t_{r,l}$} & feature vector associated with\\
      & part-of-speech~$c$ &
      & word~$S_{r,l}$ at frame~$t$ of video~$D_r$\\
      \red{$x_{r,l}$} & $(x^1_{r,l},\ldots,x^{T_r}_{r,l})$ &
      \red{$x_r$} & $(x_{r,1},\ldots,x_{r,L_r})$\\
      \green{$a^m_{0,k}$} & initial probability at state~$k$ of the HMM &
      \green{$a^m_{i,k}$} & transition probability from state~$i$\\
      & for entry~$m$, with $1 \le k \le I_{C_m}$ &
      & to state~$k$ of the HMM for entry~$m$,\\
      &&& with $1 \le i,k \le I_{C_m}$\\
      \green{$b^m_{i,n}(x)$} & output probability of observing~$x$ as the &
      $Z_{c,n}$ & number of bins for the $n$th feature\\
      & $n$th feature at state~$i$ of the HMM for &
      & of the HMM for part-of-speech~$c$\\
      & entry~$m$, with $1 \le i \le I_{C_m}$, &\\
      & $1 \le n \le N_{C_m}$, and $1 \le x \le Z_{C_m,n}$ &\\
      \green{$\lambda$} & entire HMM parameter space &
      $G_r$ & size of the competition set for video~$D_r$\\
      $S^g_r$ & $g$th sentence in the competition set for &
      $L_{r,g}$ & number of words in sentence~$S^g_r$\\
      &  video~$D_r$ &&\\
    \end{tabular}
  \end{center}
  \caption{Notation for the quantities in our formulation.
    Quantities in black are provided.
    Quantities in \red{red} are hidden.
    Quantities in \green{green} are learned.}
  \label{tab:notation}
\end{table}

An unknown \emph{participant-to-track mapping} bridges the gap between the
sentence and the video.
Consider a potential mapping $p_0 \mapsto \tau_{947}$, $p_1 \mapsto
\tau_{319}$, and $p_2 \mapsto \tau_{239}$.
This would result in the above sentence being grounded in a set of tracks as
follows:
\begin{equation*}
\textit{person}(\tau_{947})\wedge
\textit{to-the-left-of}(\tau_{947},\tau_{319})\wedge
\textit{backpack}(\tau_{319})\wedge
\textit{approached}(\tau_{947},\tau_{239})\wedge
\textit{trash-can}(\tau_{239})
\end{equation*}
In such grounding, tracks are bound to words first through the
participant-to-track mapping and then through the argument-to-participant
mapping.
This allows the HMM for each word in the sentence to be \emph{instantiated}
for a collection of tracks.
With known HMM parameters, an instantiated HMM can be used to score the
observation of features calculated from those tracks.
A sentence score can then be computed by aggregating the scores of all of the
words in that sentence.

The above mechanism can either compute a MAP estimate of the most probable
participant-to-track mapping or an exhaustive score summing all possible
such mappings.
The former can be computed with the \citet{Viterbi1967} algorithm and the
latter can be computed with the forward algorithm \citep{Baum1966}.
These computations are similar, differing only by replacing $\max$ with $\sum$.

The ML formulation scores a video-sentence pair~$r$ with:
\begin{equation}
  \label{eq:score}
  L(D_r;S_r,\lambda)=\sum_{j_r}\underbrace{P(j_r|D_r)}_{\textbf{\emph{Track}}}
  \underbrace{P(x_r|S_r,\lambda)}_{\textbf{\emph{Sentential}}}
\end{equation}
where~$j_r$ denotes a transposition of a collection of object tracks for
video clip~$r$, one per participant.
For example, if the tracks for the two participants were
$\tau_{r,239}=(4,2,7,8,3)$ and $\tau_{r,947}=(1,3,7,4,9)$ (where each element
in a sequence is the index of a detection in a particular frame, \eg\ `2' means
the second detection from the detection pool in the second frame, `7' means the
seventh detection in the third frame, \etc), then
$j_r=((4,1),(2,3),(7,7),(8,4),(3,9))$.
The sequence of features~$x_r$ are computed from tracks~$j_r$ that are bound to
the words in~$S_r$.
Eq.~\ref{eq:score} sums over the unknown participant-to-track mappings and in
each such mapping it combines a \textbf{\emph{Sentential}} score, in the form
of the joint HMM likelihoods, with a \textbf{\emph{Track}} score, which
internally measures both detection quality in every frame and temporal
coherence between every two adjacent frames.
The \textbf{\emph{Sentential}} score is itself
\begin{equation*}
  P(x_r|S_r,\lambda)=\sum_{q_r}\prod^{L_r}_{l=1} P(x_{r,l},q_{r,l}|S_{r,l},\lambda)
\end{equation*}
\begin{equation*}
  P(x_{r,l},q_{r,l}|S_{r,l},\lambda)
  =\prod^{T_r}_{t=1}a^{S_{r,l}}_{q^{t-1}_{r,l},q^t_{r,l}}
  \prod^{N_{C_{S_{r,l}}}}_{n=1}b^{S_{r,l}}_{q^t_{r,l},n}(x^t_{r,l,n})
\end{equation*}

A lexicon is learned by determining the unknown HMM parameters that best
explain the~$R$ training samples.
The ML approach does this by finding the optimal parameters~$\lambda^*$  that
maximize a joint score
\begin{equation}
  \label{eq:ml}
  L(D;S,\lambda)=\prod^R_{r=1} L(D_r;S_r,\lambda)
\end{equation}
Once~$\lambda$ is learned, one can determine whether a given video depicts a
given sentence by thresholding the score for that pair produced by
Eq.~\ref{eq:score}.

\section{Discriminative Training}
\label{section:dt}
The ML framework employs occurrence counting via Baum Welch
\citep{BaumPSW70,Baum72} on video clips paired with positive sentences.
We extend this framework to support DT on video clips paired with both positive
and negative sentences.
As we will show by way of experiments in Section~\ref{section:experiment},
DT usually outperforms ML when there is a limited quantity of positive-labeled
video clips.

Towards this end, for training sample~$r$, let~$G_r$ be the size of its
\emph{competition set}, a set formed by pooling one positive sentence and
multiple negative sentences with video clip~$D_r$.
One can extend the ML score from Eq.~\ref{eq:score} to yield a
\emph{discrimination score} between the positive sentences and the
corresponding competition sets for each training sample, aggregated over the
training set.
\begin{equation}
  \label{eq:objective1}
  O(\lambda )= \sum_{r=1}^{R}{}\left[\underbrace{\log L(D_r; S_r,
    \lambda)}_{\textbf{\emph{Positive score}}}-
  \underbrace{\log\sum_{g=1}^{G_r} L(D_r; S_r^g,
    \lambda)}_{\textbf{\emph{Competition score}}}\right]
\end{equation}
The \textbf{\emph{Positive score}} is the $\log$ of Eq.~\ref{eq:score} so the
left half of $O(\lambda)$ is the $\log$ of the ML objective function
Eq.~\ref{eq:ml}.
The \textbf{\emph{Competition score}} is the $\log$ of the sum of scores so the
right half measures the aggregate competition within the $R$ competition sets.
With parameters that correctly characterize the word and sentential meanings in
a corpus, the positive sentences should all be true of their corresponding
video clips, and thus have high score, while the negative sentences should all
be false of their corresponding video clips, and thus have low score.
Since the scores are products of likelihoods, they are nonnegative.
Thus the \textbf{\emph{Competition score}} is always larger than the
\textbf{\emph{Positive score}} and $O(\lambda)$ is always negative.
Discrimination scores closer to zero yield positive sentences with higher score
and negative sentences with lower score.
Thus our goal is to maximize $O(\lambda)$.

This discrimination score is similar to the Maximum Mutual Information (MMI)
criterion \citep{Bahl1986} and can be maximized with the Extended Baum-Welch
(EBW) algorithm used for speech recognition \citep{He2008,Jiang2010}.
However, our discrimination score differs from that used in speech recognition
in that each sentence score $L$ is formulated on a cross product of Viterbi
lattices, incorporating both a factorial HMM of the individual lexical entry
HMMs for the words in a sentence, and tracks whose individual detections also
participate in the Markov process as hidden quantities.
One can derive the following reestimation formulas by constructing the
\emph{primary} and \emph{secondary} auxiliary functions in EBW to iteratively
maximize $O(\lambda)$:
\begin{equation}
  \label{eq:ebw}
  \begin{array}{l}
    a_{i,k}^{m} = \theta^m_i{}{\displaystyle\sum^R_{r=1}}{\mathop
      {\displaystyle\sum^{G_r}_{g=1}\displaystyle\sum^{L_{r,g}}_{l=1}}
      \limits_{\text{s.t.}\;S^g_r=m}} \displaystyle\sum^{T_r}_{t=1}\left[
    \underbrace{\frac{L(q^{t-1}_{r,l}=i,q^t_{r,l}=k,D_r;S^g_r,\lambda')}
        {L(D_r;S^g_r,\lambda')}\epsilon(r,g)}_{\Delta\xi(r,g,l,i,k,t)}+C\cdot
    {a^m_{i,k}}'\right]\\
    b^m_{i,n}(h) = \phi^m_{i,n}{}{\displaystyle\sum^R_{r=1}}{\mathop
      {\displaystyle\sum^{G_r}_{g=1}\displaystyle\sum^{L_{r,g}}_{l=1}}
      \limits_{\text{s.t.}\;S^g_{r,l}=m}} \displaystyle\sum^{T_r}_{t=1}\left[
    \underbrace{\frac{L(q^t_{r,l}=i,x^t_{r,l,n}=h,D_r;S^g_r,\lambda')}
        {L(D_r;S^g_r,\lambda')}\epsilon(r,g)}_{\Delta\gamma(r,g,l,n,i,h,t)}+C\cdot
    {b^m_{i,n}}'(h)\right]
  \end{array}
\end{equation}
In the above, the coefficients~$\theta^m_i$ and~$\phi^m_{i,n}$ are for
sum-to-one normalization, $L_{r,g}$~is the number of words in sentence~$S^g_r$,
$\epsilon(r,g)=\delta(S^g_r)-L(D_r;S^g_r,\lambda') / \sum_g L(D_r; S^g_r,
\lambda')$ with $\delta(S^g_r)=1$ iff $S^g_r=S_r$, and~${a^m_{i,k}}'$
and~${b^m_{i,n}}'(h)$ are in the parameter set~$\lambda'$ of the previous
iteration.
The damping factor~$C$ is chosen to be sufficiently large so that the
reestimated parameters are all nonnegative and $O(\lambda) \ge O(\lambda')$.
In fact, $C$ can be selected or calculated independently for each sum-to-one
distribution (\eg\ each row in the HMM transition matrix or the output
distribution at each state).
The $\Delta\xi(r,g,l,i,k,t)$ and $\Delta\gamma(r,g,l,n,i,h,t)$ in
Eq.~\ref{eq:ebw} are analogous to the occurrence statistics in the reestimation
formulas of the ML framework and can be calculated efficiently using the
Forward-Backward algorithm \citep{Baum1966}.
The difference is that they additionally encode the discrimination
$\epsilon(r,g)$ between the positive and negative sentences into the counting.

While Eq.~\ref{eq:ebw} efficiently yields a local maximum to $O(\lambda)$,
we found that, in practice, such local maxima are far worse than the global
optimum we seek.
There are two reasons for this.
First, the objective function has many shallow maxima which occur when there
are points in the parameter space, far from the correct solution, where there
is little difference between the scores of the positive and negative sentences
on individual frames.
At such points, a small domination of the the positive samples over the
negative ones in many frames, when aggregated, can easily overpower a large
domination of the negative samples over the positive ones in a few frames.
Second, the discrimination score from Eq.~\ref{eq:objective1} tends to
assign a larger score to shorter sentences.
The reason is that longer sentences tend to have greater numbers of tracks and
Eq.~\ref{eq:score} takes a product over all of the tracks and all of the
features for all of the words.

One remedy for both of these problems is to incorporate a sentence prior to the
per-frame score:
\begin{equation}
  \label{eq:new-score}
  \hat{L}(D_r,S_r;\lambda)=\left[L(D_r;S_r,\lambda)\right]^{\frac{1}{T_r}}\pi(S_r)
\end{equation}
where
\begin{equation*}
  \pi(S_r)=\exp\sum_{l=1}^{L_r} \left[E(I_{C_{S_{r,l}}})
  +\sum_{n=1}^{N_{C_{S_{r,l}}}}E(Z_{{C_{S_{r,l}}},n})\right]
\end{equation*}
In the above, $Z_{C_{S_{r,l}},n}$ is the number of bins for the $n$th feature
of the word~$S_{r,l}$ whose part of speech is $C_{S_{r,l}}$ and
$E(Y)=\log Y$ is the entropy of a uniform distribution over~$Y$
bins.
Replacing~$L$ with~$\hat{L}$ in Eq.~\ref{eq:objective1} yields a new
discrimination score:
\begin{equation*}
    \hat{O}(\lambda )= \sum^R_{r=1}\left[\log\hat{L}(D_r, S_r; \lambda)-
    \log\sum_{g=1}^{G_r} \hat{L}(D_r, S_r^g; \lambda)\right]
\end{equation*}
$\hat{O}$~is smoother than~$O$ which prevents the training process from being
trapped in shallow local maxima \citep{Jiang2010}.

Unfortunately, we know of no way to adapt the Extended Baum-Welch algorithm to
this objective function because of the exponents $1/T_r$ in
Eq.~\ref{eq:new-score}.
Fortunately, for any parameter~$\lambda_{i,j}$ in the parameter set~$\lambda$
that must obey a sum-to-one constraint $\sum_k\lambda_{i,k}=1$, there exists a
general reestimation formula using the Growth Transformation (GT) technique
\citep{Gopalakrishnan1991,Jiang2010}
\begin{equation}
  \label{eq:gt}
  \lambda_{i,j} = \frac{{\lambda'_{i,j}}\left[\displaystyle\frac{\partial\hat{O}}
      {\partial \lambda_{i,j}}\bigg|_{\lambda_{i,j}=\lambda'_{i,j}}+C_i\right]}
    {\displaystyle\sum_k\lambda'_{i,k}\left[\frac{\partial \hat{O}}
        {\partial\lambda_{i,j}} \bigg|_{\lambda_{i,j}=\lambda'_{i,k}} + C_i\right]}
\end{equation}
which guarantees that $\hat{O}(\lambda) \ge \hat{O}(\lambda')$ and that the
updated parameters are nonnegative given sufficiently large values~$C_i$ for
every~$\lambda_{i,j}$, similar to Eq.~\ref{eq:ebw}.

Two issues must be addressed to use Eq.~\ref{eq:gt}.
First, we need to compute the gradient of the objective function~$\hat{O}$.
We employ automatic differentiation (AD; \citealp{Wengert1964}),
specifically the \textsc{adol-c} package \citep{Walther2009}, which yields
accurate gradients up to machine precision.
One can speed up the gradient computation by rewriting the partial derivatives
in Eq.~\ref{eq:gt} with the chain rule as
\begin{equation*}
  \frac{\partial\hat{O}}{\partial\lambda_{i,j}}
  =\sum^R_{r=1}\left[\frac{\frac{\partial\hat{L}(D_r, S_r;\lambda)}
      {\partial\lambda_{i,j}}}{\hat{L}(D_r, S_r;\lambda)} -
    \frac{\displaystyle\sum^{G_r}_{g=1}\frac{\partial\hat{L}(D_r, S_r;\lambda)}
      {\partial\lambda_{i,j}}}{\displaystyle\sum^{G_r}_{g=1}\hat{L}(D_r,
      S_r;\lambda)}\right]
\end{equation*}
which decomposes the derivative of the entire function into the independent
derivatives of the scoring functions.
This decomposition also enables taking derivatives in parallel within a
competition set.

The second issue to be addressed is how to pick values for~$C_i$.
On one hand, $C_i$~should be sufficiently large enough to satisfy the GT
conditions (\ie\ growth and nonnegativity).
On the other hand, if it is too large, the growth step of each iteration will
be small, yielding slow convergence.
We employ an adaptive method to select~$C_i$.
Let~$y$ be the last iteration in which the objective function value increased.
We select~$C_i$ for the current iteration~$w+1$ by comparison between~$y$ and
the previous iteration~$w$:
\begin{equation}
  \label{eq:adaptive}
  C_i=\left\{
  \begin{array}{l@{\hspace*{20pt}}l}
  \max\left(0,-\min_k\displaystyle\frac{\partial\hat{O}}{\partial\lambda_{i,k}}
  \bigg|_{\lambda_{i,k}=\lambda'_{i,k}}\right)   & {w=y}\\
  \max(C'_i,\varepsilon)\cdot\chi & {w>y}
  \end{array}
  \right.
\end{equation}
where~$C'_i$ is the damping factor of the previous iteration~$w$, $\chi > 1$ is
a fixed punishment used to decrease the step size if the previous iteration
failed, and $\varepsilon > 0$ is a small value in case $C'_i=0$.
Using this strategy, our algorithm usually converges within a few dozen
iterations.

\section{Training Regimen}
\label{section:training-regimen}
Successful application of DT to our problem requires that negative sentences
in the competition set of a video clip adequately represent the negative
sentential population of that video clip.
We want to differentiate a positive sentence from as many varied negative
sentences as possible.
Otherwise we would be maximizing the discrimination between a positive label
and a only small portion of the negative population.
Poor selection of negative sentences will fail to avoid local optima.

With larger, and potentially recursive, grammars, the set of all possible
sentences can be large and even infinite.
It is thus infeasible to annotate video clips with every possible positive
sentence.
Without such annotation, it is not possible to take the set of negative
sentences as the complement of the set of positive sentences relative to the
set of all possible sentences generated by a grammar and lexicon.
Instead, we create a restricted grammar that generates a small finite subset of
the full grammar.
We then manually annotate all sentences generated by this restricted grammar
that are true of a given video clip and take the population of negative
sentences for this video clip to be the complement of that set relative to the
restricted grammar.
However, the optimization problem would be intractable if we were to use this
entire set of negative sentences, as it could be large.
Instead, we randomly sample negative sentences from this population.
Ideally, we want the size of this set to be sufficiently small to reduce
computation time but sufficiently large to be representative.

Nevertheless, it is still difficult to find a restricted grammar that both
covers the lexicon and has a sufficiently small set of possible negative
sentences so that an even smaller representative set can be selected.
Thus we adopt a two-phase regimen where we first train a subset of the lexicon
that admits a suitable restricted grammar using DT and then train the full
lexicon using ML where the initial lexicon for ML contains the output entries
for those words trained by DT.\@
Choosing a subset of the lexicon that admits a suitable restricted grammar
allows a small set of negative sentences to adequately represent the total
population of negative sentences relative to that restricted grammar and
enables DT to quickly and correctly train the words in that subset.
That subset `seeds' the subsequent larger ML problem over the entire lexicon
with the correct meanings of those words facilitating better convergence to the
correct entries for all words.
A suitable restricted grammar is one that generates sentences with just nouns
and a single verb, omitting prepositions and adverbs.
Since verbs have limited arity, and nouns simply fill argument positions in
verbs, the space of possible sentences generated by this grammar is thus
sufficiently small.

\section{Experiment}
\label{section:experiment}
To compare ML and DT on this problem, we use exactly the same
experimental setup as that in the ML framework \citep{Yu2013}.
This includes the dataset (61 videos with 159 annotated positive sentences),
the off-the-shelf object detector \citep{Felzenszwalb2010a,Felzenszwalb2010b},
the HMM configurations, the features, the three-fold cross-validation design,
the baseline methods \textsc{chance}, \textsc{blind}, and \textsc{hand}, and
the twenty-four test sentences divided into two sets~\texttt{NV} and
\texttt{ALL}.
Each test sentence, either in \texttt{NV} or in \texttt{ALL}, is paired with
every test video clip.
The trained models are used to score every video-sentence pair produced by such
according to Eq.~\ref{eq:new-score}.
Then a binary judgment is made on the pair deciding whether or not the video
clip depicts the paired sentence.
This entire process is not exactly the same on the baseline methods:
\textsc{chance} randomly classifies a video-sentence pair as a hit with
probability 0.5; \textsc{blind} only looks at the sentence but never looks at
the video, whose performance can be bounded through yielding the optimal
classification result in terms of the maximal F1 score with known groundtruth;
\textsc{hand} uses human engineering HMMs instead of trained HMMs.

As discussed in Section~\ref{section:training-regimen}, we adopt a
two-phase training regimen which discriminatively trains positive and negative
sentences that only contain nouns and a single verb in the first phase and
trains all sentences over the entire lexicon based on ML in the second phase.
In the first phase, for each positive sentence in a training sample, we
randomly select 47 sentences from the corresponding negative population and
form a competition set of size 48 by adding in the positive sentence.

We compared our two-phase learning algorithm (\textsc{dt+ml}) with the original
one-phase algorithm (\textsc{ml}).
For an apples-to-apples comparison, we also implemented a two-phase training
routine with only ML in both phases (\textsc{ml+ml}), \ie\ DT in the first
phase of our algorithm is replaced by ML.\@
In the following, we report experimental results for all three
algorithms: \textsc{ml}, \textsc{ml+ml}, and \textsc{dt+ml}.
Together with the three baselines above, in total we have six methods for
comparison.

\begin{figure}[t]
  \begin{center}
    \subfigure[]{
      \includegraphics[width=0.45\columnwidth]{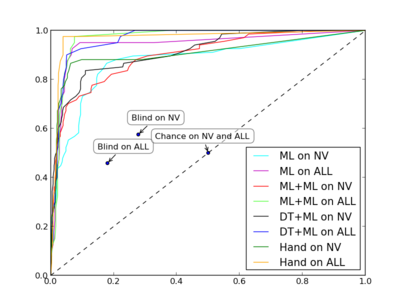}
      \label{fig:ROC-0.67}
    } \\
    \subfigure[]{
      \includegraphics[width=0.45\columnwidth]{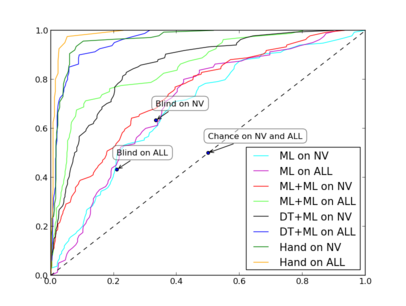}
      \label{fig:ROC-0.33}
    }
    \subfigure[]{
      \includegraphics[width=0.45\columnwidth]{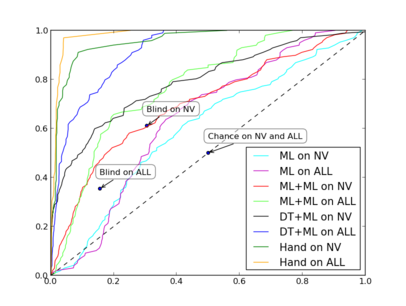}
      \label{fig:ROC-0.17}
    }
    \caption{ROC curves for size ratios of (a)~0.67, (b)~0.33, and (c)~0.17.}
    \label{fig:ROC}
  \end{center}
\end{figure}

To demonstrate the advantage of DT over ML on small training sets, we evaluate
on three distinct ratios of the size of the training set to that of the whole
dataset: 0.67, 0.33, and 0.17.
This results in about 40, 20, or 10 training video clips, tested on
the remaining 20, 40, or 50 video clips for the above size ratios
respectively.
The training and testing routines were unchanged across ratios.
We perform a separate three-fold cross validation for each ratio and then pool
the results to obtain ROC curves for that ratio.
Since \textsc{chance} and \textsc{blind} directly output a binary judgment
instead of a score on each testing video-sentence pair, the ROC curves contain
points for these baselines instead of curves.
The performance of the six methods on different ratios is illustrated in
Figure~\ref{fig:ROC}.

Several observations can be made from the figure.
First, the performance of both DT and ML gradually increases as the ratio
increases.
Their performance is far from that of \textsc{hand} on the smallest training
set with ratio 0.17 while it is very close on the largest training set
with ratio 0.67.
This implies that as the learning problem is better constrained given more
training data, both training algorithms find better local maxima.
Second, the performance gap between DT and ML gradually decreases as the ratio
increases.
With ratio 0.17, although both DT and ML perform poorly, the gap between them
is the largest.
In this case, the learning problem is highly unconstrained, which makes ML
suffer more severely from incorrect local optima than DT.\@
However, with ratio 0.67, the problem is well constrained and there is almost
no performance gap; sometimes ML performs even better than DT.\@
Third, the two-phase \textsc{ml+ml} performs generally better than the one-phase
\textsc{ml}.
Fourth, results on \texttt{ALL} are generally better than those on \texttt{NV}.
The reason is that longer sentences with varied parts of speech incorporate
more information into the scoring function from Eq.~\ref{eq:new-score}.

\section{Conclusion}
\label{section:conclusion}
We have proposed a DT framework for learning word meaning representations from
video clips paired with only sentential labels in a weakly supervised fashion.
Our method is able to automatically determine the word-to-meaning mappings from
the sentences to the video data.
Unlike the ML framework, our framework exploits not only the information of
positive sentential labels but also that of negative labels, which makes the
learning problem better constrained given the same amount of video data.
We have demonstrated that DT significantly outperforms ML on small training
datasets.
Currently, the learning problem makes several assumptions about knowing: the
grammar, the arity of each entry in the lexicon, and the participant number in
each sentence, \etc\ In the future, we seek to gradually remove these
assumptions by also learning these knowledge from training data.

\ifnipsfinal
\subsubsection*{Acknowledgments}
This research was sponsored by the Army Research Laboratory and was
accomplished under Cooperative Agreement Number W911NF-10-2-0060.
The views and conclusions contained in this document are those of the authors
and should not be interpreted as representing the official policies, either
express or implied, of the Army Research Laboratory or the U.S. Government.
The U.S. Government is authorized to reproduce and distribute reprints for
Government purposes, notwithstanding any copyright notation herein.
\fi

\begin{small}
\bibliographystyle{abbrvnat}
\setlength{\bibsep}{0.6ex}
\bibliography{arxiv2013a}
\end{small}

\end{document}